# Any-Play: An Intrinsic Augmentation for Zero-Shot Coordination

Keane Lucas
Carnegie Mellon University
Pittsburgh, PA, USA
kjlucas@andrew.cmu.edu

Ross E. Allen
MIT Lincoln Laboratory
Lexington, MA, USA
ross.allen@ll.mit.edu

## ABSTRACT

Cooperative artificial intelligence with human or superhuman proficiency in collaborative tasks stands at the frontier of machine learning research. Prior work has tended to evaluate cooperative AI performance under the restrictive paradigms of *self-play* (teams composed of agents trained together) and *cross-play* (teams of agents trained independently but using the same algorithm). Recent work has indicated that AI optimized for these narrow settings may make for undesirable collaborators in the real-world. We formalize an alternative criteria for evaluating cooperative AI, referred to as *inter-algorithm cross-play*, where agents are evaluated on teaming performance with all other agents within an experiment pool with no assumption of algorithmic similarities between agents. We show that existing state-of-the-art cooperative AI algorithms, such as Other-Play and Off-Belief Learning, under-perform in this paradigm. We propose the *Any-Play* learning augmentation—a multi-agent extension of diversity-based intrinsic rewards for zero-shot coordination (ZSC)—for generalizing self-play-based algorithms to the inter-algorithm cross-play setting. We apply the Any-Play learning augmentation to the Simplified Action Decoder (SAD) and demonstrate state-of-the-art performance in the collaborative card game Hanabi.

## 1 INTRODUCTION

Superhuman artificial intelligence (AI) has become increasingly commonplace for competitive tasks such as chess [6, 29, 31], Atari [23, 29], Go [29, 30], StarCraft II [36], DotA 2 [3], and poker [4]. The success of competitive AI is due, in large part, to recent advancements in deep reinforcement learning (RL). In contrast, the domain of cooperative AI—where autonomous agents must collaborate with humans or separately-trained agents toward a shared objective—is relatively understudied within reinforcement learning [2].

Further limiting the scope of cooperative reinforcement learning research is the fact that existing work has relied on encoding domain knowledge into the training algorithm [18] and evaluating cooperative AI in the narrow problem settings of *self-play* [16] and zero-shot coordination *cross-play* [17, 18]. Self-play (SP) refers to cooperative teams composed of agents that were all trained together, often being identical copies of one another [14, 34]. Zero-shot coordination (ZSC)[1] refers to a more general setting where agents must cooperate with other agents for which they have no prior interactions. However, past work has analyzed a narrow form of ZSC, commonly referred to as "cross-play" (XP), whereby teams are composed of agents trained independently but using the same underlying algorithm. In this paper we disambiguate this narrow form of XP by referring to it as *intra-algorithm cross-play* (intra-XP).

Recent work has indicated that training cooperative AI to maximize intra-XP performance does not necessarily produce agents that perform well in more general, real-world settings where there are no guarantees that teammates share algorithmic underpinnings. Siu et al. [32] conduct a series of human-AI teaming experiments within the collaborative card game *Hanabi* using an AI teammate with state-of-the-art intra-XP Hanabi scores. Siu et al. conclude that human Hanabi players have strong adverse reactions to playing with the high-performing intra-XP agent, and strongly prefer a different AI teammate that was not optimized for intra-XP performance. This is evidence that intra-algorithm cross-play may be a poor indicator for predicting real-world performance of cooperative AI and that a more general scoring function is needed.

In this paper we make two contributions that advance the state-of-the-art for cooperative RL. First we propose a formal definition of a more comprehensive ZSC evaluation criteria, referred to as *inter-algorithm cross-play* (inter-XP), whereby an agent's teaming performance is measured as an average over a pool of teammates with no assumption of algorithmic similarities between agents. Our second contribution is an RL training augmentation, referred to as *Any-Play*, which provides an auxiliary loss function and intrinsic reward to help self-play-based agents generalize to the more challenging setting of both intra-algorithm and inter-algorithm cross-play. The Any-Play (AP) augmentation is based on the property of *diversity* from intrinsic reinforcement learning [10] and does not require environment-specific implementations. Using the ZSC benchmark environment *Hanabi*, Any-Play agents produce near-state-of-the-art performance in self-play and intra-XP settings, and outperform all other agents in the inter-XP setting.

## 2 RELATED WORK

This section helps position the inter-XP scoring function and Any-Play learning augmentation in the context of broader research on multi-agent reinforcement learning (MARL) and zero-shot coordination (ZSC). We introduce the benchmark ZSC environment of Hanabi used in on our experiments (Sec. 4) and provide related work on Hanabi AI. We also introduce related work from the field of intrinsic reinforcement learning that motivate construction of the Any-Play augmentation presented in Sec. 3.

---
[1] Also referred to as *ad-hoc cooperation* [7]

## 2.1 Zero-Shot Coordination

Reinforcement learning (RL) and multi-agent reinforcement learning (MARL) has recently been used to create superhuman AI in single-agent [23], 1-vs-1 [29, 30, 36], 1-vs-N [4], and N-vs-N [3] competitive games. However, a relatively under-explored branch of MARL research focuses on cooperative games—such as Hanabi [2, 17, 18] and Overcooked [9, 33, 39]—and mixed cooperative/competitive games such as Pommerman [28], Bridge [41], and Diplomacy [13]. Cooperative games offer unique challenges beyond purely competitive games. Cooperative agents must often possess a *theory of mind*—i.e. the ability to infer teammates' mental states and predict their future actions—in order to optimally coordinate on shared objectives [26, 27]. Furthermore, due to the shared reward signal, it can be very difficult to even evaluate an individual agent's contribution to a collaborative game [1, 11, 40].

In this paper we focus on cooperative games in the zero-shot coordination (ZSC) setting [18]. The ZSC setting evaluates the performance of an agent in a cooperative game by pairing it with never-before-seen teammates and evaluating the resulting overall team performance. Many of the most-cited MARL algorithms perform poorly in the ZSC setting due to their reliance on *self-training*[2] and *parameter sharing* [11, 12, 14, 16, 20, 21, 34]. These techniques often lead to "secretive conventions" where agents who are trained together adopt idiosyncratic behaviors that are not interpretable by human teammates or other agents not present during training.

Hu et al. [18] propose the *Other-Play* (OP) algorithm that attempts to avoid such secretive conventions by teaming agents with "symmetrically re-labeled" copies of themselves during training. Symmetries in an environment are defined as re-labeling of states and actions that leave trajectories unchanged aside from the re-labeling itself. The OP algorithm requires the environment-specific symmetries to be given as input, which make it impractical for more complex environments where symmetries are not easily pre-determined. Bullard et al. [5] propose the Quasi-Equivalence Discovery (QED) algorithm that can iteratively discover symmetries in complex environments.

Neither OP or QED fully prevent the creation of secretive conventions because not all conventions break symmetries in the environment. To this end, the Off-Belief Learning (OBL) algorithm has been proposed by Hu et al. [17]. Instead of a symmetry-based approach, OBL proves convergence to a unique policy in ZSC settings by assuming prior actions were taken by a fixed random policy but future actions will be taken by the policy in training. This process can then be iterated by using the trained policy from one iteration as the fixed policy (replacing the random policy) in the next iteration in order to train a new, higher-level policy. While OBL achieves state-of-the-art intra-XP performance, we demonstrate in Sec. 4 that it can perform poorly in the inter-XP setting when paired with teammates not trained with OBL. This is likely due to the fact that the unique policy OBL agents converge upon has no guarantees of being robust to teaming with other policies, and non-OBL agents are unlikely to have converged upon the unique OBL policy, independently.

## 2.2 Hanabi AI

The card game *Hanabi* has become a benchmark environment for training and evaluating cooperative, zero-shot AI agents; much in the same way that chess and Go have been benchmarks for competitive AI [2]. Hanabi is akin to a multiplayer version of Solitaire where players must work together to arrange their cards in ordered stacks based on color. The game can be played with 2-5 players. Players do not see their own cards, only those of their teammates, and there are strict rules on what can be communicated between players. The game ends when all of the cards are successfully stacked or after 3 invalid actions are taken by the team. The game is scored based on the number of cards stacked at the end of the game with a maximum score of 25.

In 2018 and 2019 the Hanabi Challenge was held to instigate the development of Hanabi AI [37]. Agents were scored in two different competitive tracks, "Mirror" (i.e. self-play) and "Mixed" (i.e. inter-algorithm cross-play), with competition-winning scores of 20.6/25 and 13.2/25 in each track, respectively. This Mixed track gives a historical example of the inter-XP setting that we formalize as a general-purpose ZSC evaluation criteria in Sec. 3.1.

Siu et al. [32] provide a recent survey of performance of cooperative AI in 2-player Hanabi. Early work in Hanabi AI focused on developing rule-based agents that encoded expert knowledge of the game into decision policies [25, 35, 38]. More recent work has leveraged reinforcement learning to produce state-of-the-art performance in the SP and intra-XP settings of 2-player Hanabi. Hu and Foerster [16] propose the Simplified Action Decoder (SAD) that outperformed all of the 2018 and 2019 Hanabi Challenge competitors in the SP setting with a score of 24.0/25. SAD leveraged privileged information about a teammate's actions during a centralized training phase. It also provided a environment-specific auxiliary learning task (SAD+AUX) that aimed to predict unobserved information from the agent's observation history. The SAD self-play performance was later improved to a score of 24.6/25 by adding a search algorithm, referred to as SPARTA, on top of the learning algorithm [20]. The SAD and SAD+AUX models form the basis on which we demonstrate the Any-Play learning augmentation presented in Sec. 3. When applied to 2-player Hanabi, the previously discussed OP and OBL algorithms produced state-of-the-art performance in the intra-XP setting with top scores of 21.7/25 and 23.8/25, respectively [17, 18].

## 2.3 Diversity and Intrinsic RL

The Any-Play learning augmentation is inspired by work in the field of intrinsically-motivated reinforcement learning where agents learn useful skills within environments that lack extrinsic, "goal-oriented" reward signals. In the absence of extrinsic reward signals, intrinsic-RL agents seek to maximize information-theoretic values such as mutual information between states and actions [24] or entropy of state visitation distribution [15]. The "Diversity Is All You Need" (DIAYN) method seeks to maximize the diversity of an agent's skills (see Sec. 3.2 for definitions of "skills" and the diversity objective) [10]. Typically such intrinsically-motivated training has been used as a "warm-start" in single-agent RL that enables agents to quickly learn to solve extrinsically-motivated tasks by leveraging skills learned during the intrinsic pre-training. We adapt the DIAYN method to a multi-agent setting and use it to train diverse play-styles in one agent, referred to as the *specializer*, as well as accommodating play-styles in a partner agent, referred to as the *accommodator*. By designing the accommodator to be robust by playing with a diverse set of teammate play-styles during training, we are able to demonstrate state-of-the-art inter-XP performance when paired with a range of non-Any-Play agents in a ZSC setting.

---

[2]Also referred to as "self-play" which is closely related to—but not completely synonymous with—the self-play (SP) evaluation criteria. For example, a scripted or random bot can be evaluated in the SP setting, without ever undergoing self-play training.

The Trajectory Diversity (TrajeDi) method—published concurrently with the writing of this paper—proposes a similar, diversity-based approach to ZSC and demonstrates intra-XP performance beyond that of OP and OBL in a *modified* version of Hanabi [22]. TrajeDi forms a population of SP-trained policies that are regularized to be as diverse as possible and then trains a final policy that provides the best-response to all other policies in that population. The TrajeDi results only report a narrow set of SP and intra-XP scores; there is no consideration of inter-XP performance. Qualitatively, TrajeDi represents a much more complicated approach than Any-Play; so much so that, for it to be tested in Hanabi, the rules of the game had to be modified to make TrajeDi computationally tractable. No such environment modifications are required for the Any-Play method. Due to the timing of the publication of the TrajeDi paper, as well as the fact that it required an ad hoc variation of Hanabi, we do not provide a quantitative comparison with Any-Play in Sec. 4.3.

Fictitious co-play (FCP)—another method published while this paper was under review—also proposes a diversity-based architecture for training cooperative AI [33]. Similar to Any-Play and TrajeDi, FCP generates a pool of independently-trained collaborative agents; checkpointing the agent pool at various skill levels throughout the training process (this is akin to the Any-Play specializer agent). Then the FCP trains a "best-response" agent to all checkpointed-agents in the training pool (this is akin to the Any-Play accommodator agent). Like Any-Play, FCP proposes a much more computationally tractable approach than TrajeDi. The key shortcoming of FCP is that it relies on a behaviorally diverse training pool, yet it has no mechanism to quantify or ensure such diversity; this fact is explicitly noted in the FCP "Limitations and future work" section [33]. In contrast, we formulate Any-Play with an intrinsic augmentation that regularizes the set of specializer agents to be as quantifiably diverse as possible (see Sec. 3.2).

FCP was tested in the Overcooked cooperative environment and demonstrated state-of-the-art collaboration with human teammates. Due to the algorithmic similarities with Any-Play, FCP's success in Overcooked gives us reason to expect that Any-Play will generalize beyond the environments presented in this paper while providing a stronger theoretical grounding for diversity during training. Due to the timing of publication—as well as the fact that no FCP Hanabi agents currently exist—we do not provide a quantitative comparison with Any-Play in Sec. 4.3.

Canaan et al. [7] also propose a diversity-based approach for generating AI Hanabi agents that uses genetic algorithms and hard-coded Hanabi play conventions[3]. This approach relies heavily on expert-designed, domain-specific, rule-based systems that do not generalize to other environments. In contrast we pursue a environment-agnostic reinforcement learning approach that has potential to generalize beyond Hanabi.

## 3 METHODS AND MODELS

Our work considers Dec-POMDPs [19] defined as the tuple $(I, S, \mathcal{A}, O, T, R)$. $I$ is the finite set of $n$ agents. $S$ is the state space and the joint state of the system is $s \in S$. $\mathcal{A}$ is the joint action space and a joint action is given as $a = (a_1, a_2, ..., a_n) \in \mathcal{A}$. $O(s)$ is the joint observation function and a joint observation is given as $o = (o_1, o_2, ..., o_n) \sim O(s)$. The state transition function, $T(s'|s, a)$, represents probability of arriving in joint state $s'$ when taking joint action $a$ in state $s$. The reward is drawn from the reward function $r \sim R(s, a)$. The joint stochastic policy is represented as $\pi(a|o) = (\pi_1(a_1|o_1), ..., \pi_n(a_n|o_n))$, where $\pi_i$ represents the *local policy* of agent-$i$. A joint observation-action trajectory up to time $t$ is defined as $\tau = \left(o^{(0)}, a^{(0)}, o^{(1)}, a^{(1)}, ...o^{(t)}\right)$. The undiscounted return over trajectory $\tau$ is $G_\tau = \sum_{l=0}^{t} r^{(l)}$ and the expected return over joint policy $\pi$ is $V(\pi) = V(\pi_1, ..., \pi_n) = \mathbb{E}_{\tau \sim \pi}[G_\tau]$. Let $\alpha$ represent a stochastic *policy-generator algorithm* that generates a local policy $\pi$ from distribution $\Pi_\alpha$ (e.g. a reinforcement learning algorithm). Let $C$ represent the set of all policy generator algorithms such that $\alpha \in C$. Note that we use bold font to represent joint quantities, subscripts to represent different agents, and parenthetical superscripts to represent specific time steps.

### 3.1 Cooperative Evaluation Functions

Here we provide function definitions for scoring the performance of a policy-generator algorithm (e.g. reinforcement learning algorithm), $J(\alpha) : C \to \mathbb{R}$, under four different evaluation paradigms. For simplicity we provide 2-player forms of the evaluation functions.

**Self-Play** (SP) scores represent the expected return of any agent $i$ generated (i.e. trained) by algorithm $\alpha$ when playing cooperatively with a copy of itself.

$$J_{\text{SP}}(\alpha) = \mathbb{E}_{\pi_i \sim \Pi_\alpha}[V(\pi_i, \pi_i)] \quad (1)$$

**Intra-Algorithm Cross-Play** (intra-XP) scores represent the expected return of any agent $i$ generated by algorithm $\alpha$ when partnered with any agent $j$ that has been generated *independently* by the same algorithm $\alpha$.

$$J_{\text{intra-XP}}(\alpha) = \mathbb{E}_{\pi_i \sim \Pi_\alpha, \pi_j \sim \Pi_\alpha}[V(\pi_i, \pi_j)] \quad (2)$$

This score has been used in recent work as the default measure of cross-play performance [17, 18, 22], but excludes information about how well agents trained with a particular algorithm work with agents trained with other algorithms.

**Inter-Algorithm Cross-Play** (inter-XP) score is the expected return of any agent $i$ generated from algorithm $\alpha \in C$ when partnered with any other agent $j$ generated independently from a separate algorithm $\beta \in C \setminus \alpha$

$$J_{\text{inter-XP}}(\alpha) = \mathbb{E}_{\pi_i \sim \Pi_{\alpha \in C}, \pi_j \sim \Pi_{\beta \in C \setminus \alpha}}[V(\pi_i, \pi_j)] \quad (3)$$

---
[3]Many details have been omitted, please see Canaan et al. [7].

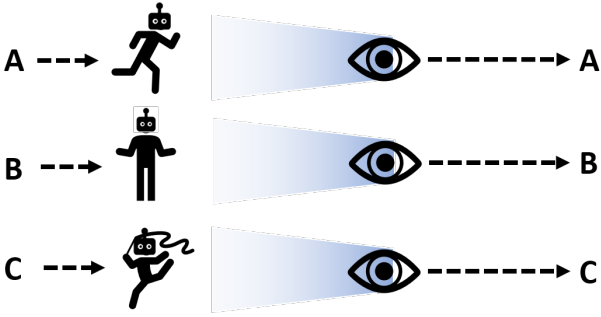

Figure 1: DIAYN - Diverse skills are learned by an agent (left) without an external reward signal by making each skill uniquely distinguishable by a discriminator (right).

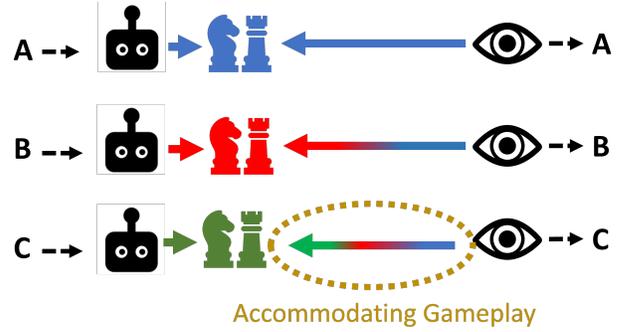

Figure 2: Any-Play - Ideal accommodative result of incorporating an intrinsic diversity reward in a cooperative MARL environment for ZSC. The accommodator (eye) learns to discern the specializer's (robot) latent variable (intent) based on its actions, in addition to learning to provide the best responding action to that action-intent pair.

Equation 3 gives a more formal definition of an evaluation paradigm that has been loosely defined in other works (e.g. the "Mixed" track in the Hanabi Challenge) [8, 9, 33, 37]. However, much of the prior work on cooperative reinforcement learning has largely ignored the inter-XP setting [17, 18, 20, 22].

Inter-algorithm cross-play scores depend on the *experiment pool*, $\hat{C}$, of other algorithms used as teammates. This has several implications. First, inter-XP scores are not relevant when compared across *different* experiment pools. Second, it is possible to bias an experiment pool to favor a specific algorithm or class of algorithms. For example, if we have an experiment pool composed largely of algorithms designed for zero-shot coordination (e.g. OP, OBL, AP), then these algorithms *may* have an unfair advantage in the inter-XP settings compared to algorithms in the pool that weren't explicitly built on the ZSC premise. Canaan et al. [7] propose a way for systematically generating experiment pools from a narrow class of rule-based agents; however, it does not allow for generating pools of modern learning-based agents [17, 18]. Therefore we need a way to evaluate the inter-XP performance between state-of-the-art ZSC algorithms while accounting for potential bias in the experiment pool toward ZSC algorithms.

**One-sided-ZSC Cross-Play** (1SZSC-XP) setting is an attempt to balance this trade-off between bias and comparison with the most relevant learning algorithms. Let $\mathcal{B}$ represent the set of algorithms that were not explicitly designed for the zero-shot coordination setting (in our experiments in Sec. 4 we have $\mathcal{B} = \{\text{IQL, VDN, SAD}\}$). The 1SZSC-XP score represents the expected score of any agent $i$ from a ZSC-based algorithm $\alpha \in C \setminus \mathcal{B}$ when partnered with any agent $j$ trained with non-ZSC algorithm $\beta \in \mathcal{B}$.

$$J_{1\text{SZSC-XP}}(\alpha) = \mathbb{E}_{\pi_i \sim \Pi_{\alpha \in C \setminus \mathcal{B}}, \pi_j \sim \Pi_{\beta \in \mathcal{B}}} \left[ V(\pi_i, \pi_j) \right] \quad (4)$$

### 3.2 Extending DIAYN for ZSC

In order to maximize inter-XP score our augmentation, referred to as **Any-Play**, attempts to regularize agent behavior to be more 'accommodative' to the many possible different ways of succeeding in a cooperative environment. The method we use is inspired by DIAYN's [10] learning of distinct *skills* without an external reward.

*DIAYN Objective.* To describe the DIAYN objective, we borrow their notation, where $S$ and $A$ are random variables representing states and actions, respectively. $Z \sim p(z)$ is a latent variable on which an agent's policy is conditioned, and a policy conditioned on a fixed $Z$ is the **skill** referred to earlier [10]. $I$ and $\mathcal{H}$ represent *mutual information* and *Shannon entropy*. In the DIAYN objective, the entropy of actions given states $\mathcal{H}[A|S]$ and the mutual information between states and skills, $I(S; Z)$ is maximized. This incentivizes agents to behave randomly, but also encodes influence of the sampled $Z$ on which states are visited for a policy conditioned on $Z$. To ensure that states visited, and not actions, distinguish the skills from each-other, the mutual information between skills and actions given the state, $I(A; Z|S)$ is minimized. The DIAYN objective is then:

$$\mathcal{F}(\theta) = I(S; Z) + \mathcal{H}[A|S] - I(A; Z|S) \quad (5)$$

DIAYN then rearranges this objective to be $\mathcal{F} = \mathcal{H}[Z] - \mathcal{H}[Z|S] + \mathcal{H}[A|S, Z]$ which implies $Z$ must have high entropy, but also implies that $Z$ should be inferrable from $S$, which is approximated via a learned discriminator $q_\psi(z|s)$ making the new objective to maximize (again, quoting from DIAYN [10]):

$$\mathcal{F}(\theta) = \mathcal{H}[A|S, Z] + \mathbb{E}_{z \sim p(z), s \sim \pi(z)}[\log(q_\psi(z|s)) - \log(p(z))] \quad (6)$$

where $\pi(z)$ is a policy conditioned on $z$ (a skill). For the full reasoning and derivation, please refer to Eysenbach et al. [10].

The implementation of this objective is accomplished by iterating an unsupervised exploration stage and a supervised discrimination stage. The exploration stage is accomplished by training a soft actor critic to learn a policy conditioned on the the uniformly sampled latent variable $z$ attempting to maximize the intrinsic reward $r_z(s, a) = \log(q_\psi(z|s)) - \log(p(z))$. The discrimination stage updates $q_\psi(z|s)$ to better discriminate between skills (policies conditioned on differing fixed $z \in Z$).

The end result of training a policy and a discriminator in this way is represented abstractly in Figure 1, where the policy must learn to act in random but distinct ways depending on its input for the discriminator $q_\psi$ to be able to distinguish the skill it is performing.

*Any-Play Objective.* We use this objective of learning distinct skills from random navigation of an MDP to augment the cooperative MARL self-play objective of maximizing an external environment-specific (extrinsic) reward. Also, because we consider environments that are partially observable, we replace $s$ with $o$ where $o$ is an agent's observation. Differing from DIAYN, we refer to the latent random variable $Z$ in this paper as *intent* and the policy conditioned on a fixed $z$ as a *play-style* (instead of a skill). This change in terminology is due to the external reward encoding the need for all policies conditioned on a fixed $z$ (play-styles / skills) to both maximize the external environment reward (get a high score), but still be distinguishable from each-other by a discriminator. This suggests that each policy resulting from a specific intent ($z$), will be a distinct winning strategy, or play-style. The key idea behind leveraging this objective for ZSC is that an agent trained with a partner exhibiting multiple distinct play-styles will learn a policy that is *accommodative* of many different ways of accomplishing the cooperative goal of the environment. This idea is represented in Figure 2 as a multi-colored line resulting from the discriminator partnering with multiple different play-styles.

To augment a multi-agent system with Any-Play, we need to add the DIAYN components (the conditioned policy and the intent discriminator). We condition one of the agents on a sampled latent variable $z$, and train a discriminator $q_\psi$ to distinguish the resultant different play-styles during training. The ability of the discriminator to correctly distinguish play-styles composes the intrinsic reward, which is scaled appropriately and added to the external environment specific reward.

$$R_{\text{Any-Play}}(s, o, a, z) = R_{\text{env}}(s, a) + \lambda \log(q_\psi(z|o)) \quad (7)$$

where $\lambda$ is a scaling hyper-parameter.

*Augmenting SAD Hanabi Agents with Any-Play.* In the two-player Hanabi scenario, we designate one player as the **specializer** and the other as the **accommodator**. We use a uniform categorical random distribution with $N$ possible intents to sample our intent $z$ at the beginning of every episode $Z \sim \{1, ..., N\}$. During training, the specializer is given $z$ as a one-hotted vector of length $N$, as shown in Figure 3. For the intent discriminating component of Any-Play, a dense neural net $q$ composed of the accommodator's LSTM and a single-layer DNN is added with the supervised task of correctly predicting the sampled $z$ throughout the episode. We use categorical cross-entropy as the loss for the discriminator $q$:

$$L_q = -\langle z, log(\hat{z})\rangle \text{ where } \hat{z} = q_\psi(o) \equiv [q_\psi(1|o), ..., q_\psi(N|o)] \quad (8)$$

where $z$ is a one-hot vector and $\langle \cdot, \cdot \rangle$ is the inner product. This loss is scaled by $\lambda$ and is back-propagated through the LSTM and lower-level components of the accommodator SAD architecture during training. Reward is shared between the accommodator and specializer and consists of the externally given Hanabi reward (number of cards correctly played) added to the intrinsic intent discrimination reward, which is the negated and scaled intent loss.

$$R_{\text{Any-Play},t}(s_t, o_t, a_t, z) = R_{\text{Hanabi}}(s_t, a_t) + \lambda \langle z, \log(q_\psi(o_t))\rangle \quad (9)$$

**Algorithm 1:** Any-Play training on a 2-player game

**Input** : $N$, env, numGames
**Output:** $\pi_{\text{spec}}, \pi_{\text{accomm}}$

1 **for** $i \in \{0, ..., numGames\}$ **do**
2     $z \sim RandomInt(0, N)$ ;
3     $o \leftarrow env.reset()$;
4     **while** not $o.gameEnded$ **do**
5        $a_{\text{spec}} \leftarrow \pi_{\text{spec}}(z, o)$;
6        $a_{\text{accomm}} \leftarrow \pi_{\text{accomm}}(o)$;
7        $\hat{z} \leftarrow q_\psi(o)$;
8        $l_z \leftarrow -\langle z, log(\hat{z})\rangle$;
9        $o, r \leftarrow env.step(a_{\text{spec}}, a_{\text{accomm}})$;
10       $\psi \leftarrow \psi - \eta \frac{\partial l_z}{\partial o}$;
11       $r \leftarrow r + \lambda l_z$ ;
12       update $\pi_{\text{spec}}, \pi_{\text{accomm}}$
13     **end**
14 **end**
15 **return** $\pi_{\text{spec}}, \pi_{\text{accomm}}$;

Intuitively, the reward is higher when the accommodator can correctly guess the intent of the specializer based on its play-style. Pseudo-code for this training process can be seen in Alg 1. The optimal value for $\lambda$ varies based on the number of intents $N$ and agent architecture. To find a suitable $\lambda$, we use heuristics that restart training and modify $\lambda$ if intent loss does not decrease ($\lambda$ too low) or the reward does not increase ($\lambda$ too high) within the first several training epochs at the beginning of experimentation.

## 4 EXPERIMENTS

We first show how Any-Play augmentation can achieve the perfect ZSC policy on a simple referential game taken from Hu et al. [17]. We then survey Any-Play results on the more complex ZSC benchmark Hanabi. Our code for running the Hanabi experiments is publicly available at https://github.com/mit-ll/hanabi_AnyPlay which is a fork of the SAD and OP codebase [16, 18].

### 4.1 Any-Play in a Simple Environment

The game consists of two players: Player 1 and Player 2. After observing one of two randomly selected objects ("cat" or "dog"), Player 1 can decide to leave the game for a small reward of 1, send a message to Player 2 consisting of either a 0 or 1, or incur a penalty of 5 by "lifting the curtain" to reveal the true object to Player 2. After Player 1's action, Player 2 can either leave the game for a small reward of 0.5 or guess "cat" or "dog". If Player 2's guess is correct, then both players receive a large reward of 10; an incorrect guess penalizes both agents with a reward of −10.

In a purely self-play setting, the optimal strategy for Player 1 and Player 2 is to arbitrarily agree on a mapping (i.e. idiosyncratic convention) between Player 1's observation of the object and the action of sending 0 or 1, leading to a score of 10 for every SP game. In some training instances, sending 1 could come to mean "cat", but in others it could come to mean "dog". In a ZSC setting, however, the most robust policy is for Player 1 to always lift the curtain and for Player 2 to only guess the object identity when the curtain is lifted since Player 2 has no way of knowing what convention the independently-trained Player 1 has adopted for its message passing. This leads to a best-case ZSC score of 5.

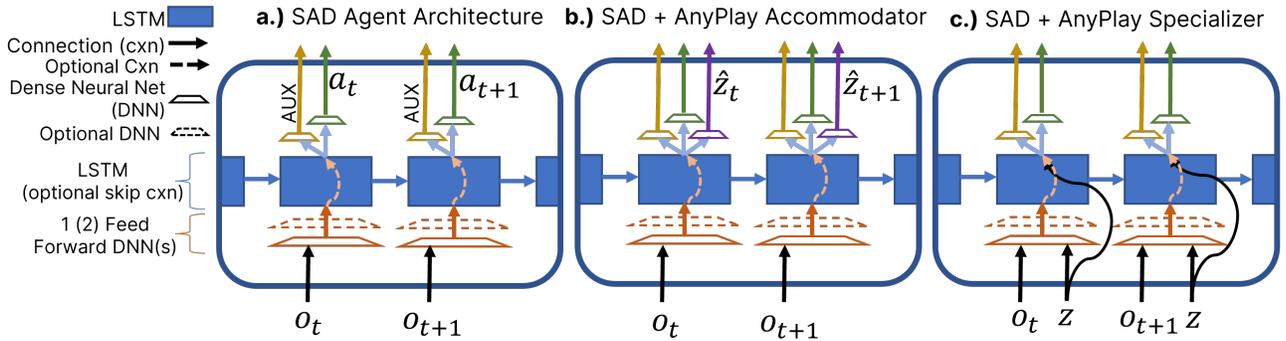

Figure 3: a.) Baseline Architecture (SAD). An optional second DNN and skip connections are used to produce diverse architectures. b.) The accommodator uses the output of the LSTM to predict the intent $\hat{z}$ of the specializer. c.) The specializer uses a policy conditioned on the intent ($z$) sampled at the beginning of each game, incentivizing diverse play-styles across multiple games.

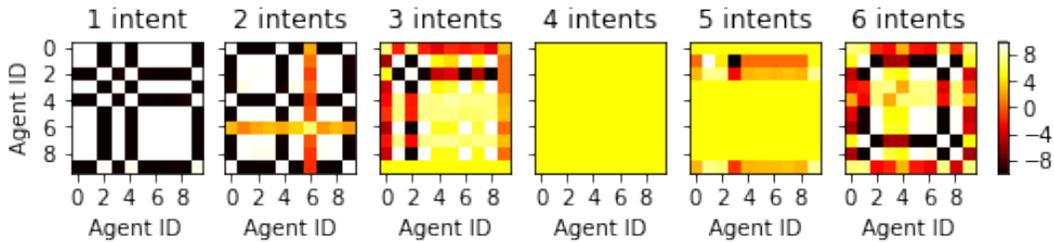

Figure 4: Simple environment cross-play matrices using Any-Play. Each matrix represents cross-play between 10 independently trained agents each augmented with Any-Play, with varying number of intents. With Any-Play using four intents, the all-yellow square indicates the best ZSC policy is achieved, where all agent pairings are compatible, matching the best-case of Off-Belief Learning [17].

Figure 4 illustrates intra-XP scores between 10 independently trained Any-Play agents with 1-6 intents for the specializer. The *1 intent* plot is a baseline that is equivalent to not augmenting the training with Any-Play, since correctly predicting one intent is trivial. Here we see a checkerboard pattern where white squares correspond to maximum-scoring pairs that occur when separate agents happen to have the same mapping for Player 1's messages and black squares are minimum-scoring pairs where agents have reversed mappings of messages.

To understand the results for the 2-6 intent cross-play matrices in Figure 4, remember that—during training—the Any-Play learning augmentation has the dual objectives of maximizing extrinsic (environment-based) reward as well as making its latent intent variable, $z$, as easy to discern as possible. Discerning latent intent is given priority over the extrinsic reward during training by scaling $\lambda$ in Equation 7. However, Player 1 only selects one of four possible actions per game (i.e. 2 bits of information per action), leaving Player 2 to discern both latent intent variable and environmental best action from a single piece of 2-bit information. After training—during ZSC evaluation—only the extrinsic reward is considered, not the intrinsic latent variable discrimination accuracy.

When Player 1's action space is larger than its latent input space (2 and 3 intents), it is possible for Player 1 to use its action to convey its latent variable (< 2 bits) as well as some information about its observation of the object. When the action space is smaller than the latent input space (5 and 6 intents), then the action cannot uniquely identify the latent variable, let alone the latent variable and the object observation. In either case, during self-play training, players can benefit from forming idiosyncratic conventions that convey more or less information about the extrinsic environment or latent variable. However these conventions can lead to imperfect ZSC scores where only the extrinsic reward matters. This is why the 2, 3, 5, and 6 intent cross-plays show a checkerboard of varying ZSC performance.

In the 4 intent case, the action space is exactly the size of the latent input space, and self-play training converges to Player 1 perfectly communicating its intent variable $z$ (due to the scaling of $\lambda$ rewarding such a policy in training), but giving no information of its observation. Therefore, with the lifted curtain being the only source of information on Player 1's observation, Player 2 learns to only trust the lifted curtain action in order to make a guess at the object. This produces the best ZSC policy and the uniform cross-play matrix shown in Figure 4.

## 4.2 Hanabi Experiment Setup

In this work, we only consider 2-player Hanabi games between AI agents. Our algorithm experiment pool, $\hat{C}$, is composed of a total of 128 different models trained from nine different policy-generator algorithms; see Table 1 for list of algorithms. We apply the Any-Play learning augmentation to the Simplified Action Decoder algorithm—referring to this combination as SAD+AP—and generate 12 different trained models using the same set of neural network architectures presented in Hu et al. [18]. Similarly we apply Any-Play to the auxiliary-task version of SAD [16] and generate 12 more trained models for experimentation, referred to as SAD+AUX+AP. For all AP-based algorithms, the accommodator is always used for cross-play evaluations; the specializer is only used during training and not involved with evaluation. We compare our 12 SAD+AP and 12 SAD+AUX+AP trained models with 104 baseline models across seven algorithms; these models have been graciously provided by authors of previous work. These baseline models include (quantity in parentheses): IQL (12), VDN (13), SAD (13), and SAD+AUX (13) from the original SAD paper [16]; the architecturally diverse SAD (12), SAD+AUX (12), SAD+OP (12), and SAD+AUX+OP (12) trained models from Other-Play [18]; and the level-5 (most competitive) trained models (5) from Off-Belief Learning (OBL) [17].

Both the SAD+AP and SAD+AUX+AP algorithms are used to train on four separate neural net architectures, with each architecture used to train three models with unique seeds, resulting in a total of 12 trained models for each algorithm. The four separate architectures consist of the four possible combinations of having one or two feed forward networks for the input, and including or not including skip connections across the recurrent network. We use the same distributed deep double Q-learning RL algorithm, prioritized replay, and dueling network architecture as SAD, Other-Play, and OBL [16–18]. Figure 3a details the neural net architecture of SAD and SAD+AUX. The models are trained on a total of 500,000 batches with each batch consisting of 128 full-game trajectories.

Through a parameter search, we found that an intent number of $N = 16$ (when there are 2 feed forward DNNs and no skip connections) and $N = 64$ (otherwise) gave consistently good results. These values are used to generate the data in Table 1. However, other numbers of intents, such as $N \in \{8, 16, 32, 48\}$ also results in intra-XP and inter-XP well above the baseline of SAD and SAD+AUX.

Algorithms are evaluated under the scoring functions described in Sec. 3.1 with each pair of agents playing 2,500 games together. For example, for the intra-XP score of SAD (from Eq. 2), all SAD agents, across all architectures, are paired against each-other for 2,500 games. The mean of all these games, excluding where the same agent plays itself, is the empirical SAD intra-XP score. For the 1SZSC-XP setting, each model from each algorithm in the set $C \setminus B$ = { SAD+AUX, SAD+OP, SAD+AUX+OP, OBL, SAD+AP, SAD+AUX+AP} is cross-played against every model in set $B$ = {IQL, VDN, SAD}.

To get the results in Table 1, we cross-played 128 different trained agents (104 from prior work and 24 of ours), resulting in 16,384 separate pairings of 2,500 games each, resulting in almost 41M games played. This large amount of cross-play games gives us our low standard error when compared to prior work [16–18].

| Algorithm | Self-Play | Intra-XP | Inter-XP | 1SZSC-XP |
|---|---|---|---|---|
| IQL | 23.8± .01 | 11.9± .01 | 11.1± .00 | N/A |
| VDN | 23.8± .01 | 8.1± .01 | 9.2± .00 | N/A |
| SAD | 23.9± .01 | 4.5± .01 | 7.3± .00 | N/A |
| SAD+AUX | 24.1± .01 | 17.7± .01 | 13.1± .00 | 5.9± .02 |
| SAD+OP | 23.9± .01 | 15.3± .01 | 12.2± .00 | 5.8± .02 |
| SAD+AUX+OP | 24.1± .01 | 22.1± .01 | 13.1± .00 | 5.6± .02 |
| OBL | **24.2± .02** | **24.2± .01** | 5.2± .00 | 1.0± .02 |
| SAD+AP | 21.6± .01 | 13.5± .01 | 10.3± .00 | 6.4± .02 |
| SAD+AUX+AP | 22.5± .02 | 20.4± .01 | **14.2± .00** | **7.4± .02** |

Table 1: Cross-play scores with standard error. While OBL outperforms in self-play and intra-XP, Any-Play augmenting SAD+AUX outperforms all other algorithms in inter-XP, with both Any-Play algorithms performing best in 1SZSC-XP. IQL, VDN, and SAD form the cross-play set for 1SZSC-XP evaluation, but are not, themselves, evaluated under this paradigm.

## 4.3 Hanabi Experimental Results

Table 1 summarizes the experimental results with nine algorithms—i.e. *policy-generators* used to train 128 distinct agents—evaluated in self-play (Eq. 1), intra-XP (Eq. 2), inter-XP (Eq. 3), and 1SZSC-XP (Eq. 4) settings. OBL scores highest in the SP and intra-XP settings, followed closely by SAD+AUX+OP. This is in line with results from prior work that show high intra-XP performance of these algorithms [17, 18] and helps validate our experimental setup. SAD+AUX+AP scores the highest in the inter-XP and 1SZSC-XP settings when compared to all other algorithms within the experiment pool listed in Table 1. This supports our claim that the Any-Play learning augmentation helps agents better collaborate in ZSC settings with never-before-seen teammates, even with those teammates that have no algorithmic similarities.

We note that OBL performs particularly poorly in the inter-XP and 1SZSC-XP settings. This seems to imply that the OBL collaboration relies heavily on shared algorithmic underpinnings, particularly the convergence to a unique policy by independently trained agents. Interestingly, the Any-Play algorithms had the lowest self-play scores[4] yet highest inter-XP scores. This is further evidence that self-play scores are poor indicator for zero-shot coordination performance.

As seen in Table 1, Any-Play (AP) significantly improves the intra-XP score of SAD from 4.5 to 13.5, and provides some improvement to SAD+AUX (17.7 → 20.4). Any-Play augmentation alone (SAD+AP) increases inter-XP of SAD by 3 points.

While SAD+OP scores better than SAD+AP in intra-XP and inter-XP, it achieves these scores by explicitly constraining the space of learnable policies by destroying information (shuffling the color of cards) deemed by the algorithm designer to likely cause conflicting behavior in Hanabi. Any-Play's effect on the otherwise non-ZSC-capable SAD's intra-XP, inter-XP, and 1SZSC-XP scores makes it comparable (and in some cases superior to) previous state-of-the-art ZSC augmentations [17, 18]. Furthermore, this effect is achieved without destroying any information during training.

---
[4]Note that these self-play scores are still better then the competition-winning SP scores from the 2018/2019 Hanabi Challenge [37].

|         | Self-Play | Intra-XP | Inter-XP | 1SZSC-XP |
|---------|-----------|----------|----------|----------|
| Self-Play | 1 | 0.055 | -0.23 | -0.55 |
| Intra-XP |   | 1 | 0.24 | -0.08 |
| Inter-XP |   |   | 1 | 0.89 |
| 1SZSC-XP |   |   |   | 1 |

Table 2: Pearson correlation coefficients between cooperative evaluation scores. 0 indicates no relationship, 1 (resp. -1) indicates perfect positive (resp. negative) correlation.

In the 1SZSC-XP setting SAD+AP outperforms SAD+OP, SAD+AUX, SAD+AUX+OP, and OBL. In addition, combining AUX with Any-Play results in an even higher 1SZSC-XP score. These scores show that, even without environmental knowledge or knowledge of the non-ZSC trained agents, you can achieve notably greater cooperation with agents that were not trained to be zero-shot cooperative.

We also conduct a correlation analysis to understand if there are aggregate trends in SP, intra-XP, inter-XP, and 1SZSC-XP scores across algorithms. Table 2 provide the Pearson correlation coefficients between pairs of scoring functions based on the data in Table 1. Interestingly, while self-play and intra-XP seem almost uncorrelated, there is a notable negative correlation between self-play and inter-XP as well as self-play and 1SZSC-XP. This may be an indication that algorithm-specific conventions (e.g. OBL's convergence to a unique policy in independent training runs) can raise self-play and/or intra-XP scores, but result in lower inter-XP and 1SZSC-XP scores. There is also a strong positive correlation between inter-XP and 1SZSC-XP scores, which may be expected due to the fact they are both forms of inter-algorithm cross-play but with differing experiment pools. We also observe a modest positive correlation between intra-XP and inter-XP scores, which is not immediately obvious from Table 1.

### 4.4 Experiment Pool Bias

Of the 104 baseline models with which we compare and cross-play our augmentation, all of them use Q-learning, 92 of them are built on top of VDN, and 74 of them incorporate the SAD augmentation on top of VDN [16, 18], as OBL excludes SAD [17]. The fact that so many of the agents in our experiment pool are based upon VDN and SAD is a potential source of experimental bias (see Sec. 3.1 for further discussion). However, these algorithmic similarities in the experiment pool are necessary if we want to draw direct comparisons with state-of-the-art Hanabi agents since many have grown from the same foundational work on SAD [16].

In investigating any potential bias these common characteristics could have in regards to unfairly helping our algorithms, SAD+AP and SAD+AUX+AP, we see that our algorithms may actually be disadvantaged by applying Any-Play on top of SAD, as opposed to applying it on IQL or VDN. This potential disadvantage comes from the fact SAD under-performs IQL and VDN in intra-XP and inter-XP settings, in spite of the large number of SAD-based models in the experiment pool. Therefore it is conceivable that we could improve our Any-Play intra-XP and inter-XP results by augmenting VDN with Any-Play, instead of SAD; and that SAD-augmentation does not give us an unfair bias in the experiment pool.

## 5 CONCLUSION

This work gives formal definition to the *inter-algorithm cross-play* evaluation function to address the shortcoming in cooperative AI research whereby agents have, historically, been evaluated in the contrived settings of self-play and intra-algorithm cross-play for zero-shot coordination. We propose *Any-Play*—an intrinsic, domain-independent training augmentation—and show how it can improve intra-algorithm cross-play and outperform state-of-the-art baseline algorithms in inter-algorithm cross-play within the card game Hanabi. Furthermore the Any-Play augmentation quantifies and regularizes diversity of agents during training, thus addressing a shortcoming of similar methods such as Fictitious Co-Play [33].

This work opens several important questions for future work, particularly relevant for human-AI teaming. Are inter-algorithm cross-play scores a better predictor of human preference in zero-shot human-AI teaming than self-play and intra-algorithm scores? If not, then what are the quantifiable objective functions on which AI can be trained in order to increase subjective human trust of AI teammates?

### ACKNOWLEDGMENTS

We would like to acknowledge the generous contribution of trained baseline models from prior work [16–18]. We would also like to thank Lujo Bauer, Jaime Pena, Ho Chit Siu, and Yutai Zhou for providing guidance on this work. This work was supported in part by a DoD National Defense Science and Engineering Graduate fellowship.

DISTRIBUTION STATEMENT A. Approved for public release. Distribution is unlimited.

This material is based upon work supported by the Under Secretary of Defense for Research and Engineering under Air Force Contract No. FA8702-15-D-0001. Any opinions, findings, conclusions or recommendations expressed in this material are those of the author(s) and do not necessarily reflect the views of the Under Secretary of Defense for Research and Engineering.